\documentclass{class_nature}

\usepackage{amsmath,amsfonts,bm}

\def\eqref#1{equation~\ref{#1}}

\def\1{\bm{1}}

\DeclareMathAlphabet{\mathsfit}{\encodingdefault}{\sfdefault}{m}{sl}
\SetMathAlphabet{\mathsfit}{bold}{\encodingdefault}{\sfdefault}{bx}{n}

\newcommand{\KL}{D_{\mathrm{KL}}}

\usepackage[latin1]{inputenc}
\usepackage[T1]{fontenc}
\usepackage{textcomp}
\usepackage{gensymb}
\usepackage{amsmath} 
\usepackage{color}
\usepackage{graphicx}
\usepackage{xfrac}
\usepackage{soul} 
\usepackage{lineno} 
\usepackage[english]{babel}
\usepackage{blindtext}
\usepackage{amssymb}
\usepackage{caption}
\usepackage{array}
\usepackage{makecell}
\usepackage{comment}
\usepackage{float}
\usepackage{hyperref}
\usepackage{url}

\graphicspath{ {./Figures/} }

\def\reals{\mathbb{R}}
\newcommand\bigzero{\makebox(0,0){\text{\large0}}}

\makeatletter
\DeclareRobustCommand\bfseriesitshape{%
  \not@math@alphabet\itshapebfseries\relax
  \fontseries\bfdefault
  \fontshape\itdefault
  \selectfont
}
\makeatother
\DeclareTextFontCommand{\textbfit}{\bfseriesitshape}

\title{Logically Synthesized, Hardware-Accelerated, Restricted Boltzmann Machines for Combinatorial Optimization and Integer Factorization} 

\author{Saavan Patel,$^{1}$ Philip Canoza,$^{1}$ Sayeef Salahuddin$^{1}$}

\begin{document}

\maketitle

\begin{affiliations}
 \item Department of Electrical Engineering and Computer Sciences, University of California, Berkeley, California 94720, USA
\end{affiliations} 



\begin{abstract}

The Restricted Boltzmann Machine (RBM) is a stochastic neural network capable of solving a variety of difficult tasks such as NP-Hard combinatorial optimization problems and integer factorization. The RBM architecture is also very compact; requiring very few weights and biases. This, along with its simple, parallelizable sampling algorithm for finding the ground state of such problems, makes the RBM amenable to hardware acceleration. However, training of the RBM on these problems can pose a significant challenge, as the training algorithm tends to fail for large problem sizes and efficient mappings can be hard to find. Here, we propose a method of combining RBMs together that avoids the need to train large problems in their full form. We also propose methods for making the RBM more hardware amenable, allowing the algorithm to be efficiently mapped to an FPGA-based accelerator. Using this accelerator, we are able to show hardware accelerated factorization of 16 bit numbers with high accuracy with a speed improvement of 10000x over a CPU implementation and 1000x over a GPU implementation, along with a power improvement of 32x and 6x, respectively.

\end{abstract}



As the demand for processing of big data accelerates, new architectures and computing paradigms are receiving heightened attention. Many of the most challenging computational problems fall into the category of NP-Hard and NP-Complete, where no exact polynomial time solution exists. For this reason, these problems would benefit the greatest from acceleration through novel architectures on FPGAs, ASICs, and new devices. \cite{Colwell2013TheLaw, Waldrop2016MoreMoore} In this context, the Ising Problem has long been known to be in the class of NP-Hard problems. Because of this, a large class of combinatorial optimization problems can be reformulated as Ising problems and solved by finding the ground state of that system \cite{Barahona1982OnModels,Kirkpatrick1983OptimizationAnnealing,Lucas2014IsingProblems}. The Boltzmann Machine \cite{Ackley1985AMachines} was originally introduced as a constraint satisfaction network based on the Ising model problem, where the weights would encode some global constraints, and stochastic units were used to escape local minima. The original Boltzmann Machine found favor as a method to solve various combinatorial optimization problems \cite{Korst1989CombinatorialMachine}. However, convergence of the sampling and training schemes on the Boltzmann Machine has been slow. \cite{Hinton2002TrainingDivergence}

To this point, the Restricted Boltzmann Machine (RBM), which is a special form of the more generic Boltzmann Machine, provides a scalable hardware architecture by eliminating intra-layer connections, while maintaining the ability to fully approximate a probability distribution over binary variables \cite{Hinton2002TrainingDivergence}. Nonetheless, even for the RBM, the convergence of the Markov Chain Monte Carlo (MCMC) algorithm, which is typically used for sampling, has proved to be challenging \cite{Tieleman2008TrainingGradient, Tieleman2009UsingDivergence}. On the other hand, the structure of the RBM lends itself to hardware acceleration. Therefore, just in the same way the continued advancement of the Moore's Law has enabled a revolution in the deep learning community, it is conceivable that hardware acceleration will make it possible to overcome the sampling problem in the RBM and allow rapid solutions of combinatorial optimization problems.     

In this work, we present an end-to-end RBM implementation that combines advances in training, model quantization and an efficient hardware implementation for inference to demonstrate substantial acceleration over standard CPU and GPU implementations. First, we propose a generative model composed of multiple learned modules that is able to solve a larger problem than the individually trained parts. This allows for circumventing the problem of training large modules, thus minimizing training time and enabling a higher degree of model accuracy. The advancement in training is accomplished through a novel merging procedure, which is based on the idea of training two models with overlapping states and then combining them along the intersection states. This is similar to combining circuits in a digital logic, with the output of one circuit sharing the same state as the input of the next. We have combined this method of training large models with algorithmic improvements that allow for compressed weights and therefore enables the use of lower precision representation to develop an efficient FPGA based accelerator. Altogether, the RBM shows approximately a $10^4$X speed acceleration and 32X power improvement for a 16 bit integer factorization problem, which is equivalent to solving a $2^{32}$ phase space, in comparison to implementing the same algorithm in the CPU or GPU.

\section*{Algorithm Details}
\label{sec:alg}
~
The RBM is a binary stochastic neural network, which  can be understood as a Markov random field of Bernoulli random variables divided into a bipartite graph structure, with the two layers called the visible states and the hidden states, graphically demonstrated in Figure \ref{fig:alg} A). We denote $v$ as the visible state, $h$ as the hidden states, and $E(v, h)$ as the energy associated with those states. The probability assigned to a given state $p(v, h) = \frac{1}{Z} e^{-E(v, h)}$ where $Z = \sum_{v, h} e^{-E(v, h)}$ is the normalizing constant of the distribution. The weight matrix $W$ and biases $a$ and $b$ create connections between the hidden and visible layers and create the probability distribution.

The bipartite structure means the visible and hidden state can be factored as $p(v | h) = \prod_i p(v_i|h)$ and $p(h | v_i) = \prod_j p(h_j|v)$ due to the conditional independence of states within the same layer. Sampling on the RBM is performed via Block Gibbs Sampling \cite{Hinton2002TrainingDivergence, Geman1987StochasticImages}, where the units in each layer are sampled in parallel.  In Figure \ref{fig:alg} B) we show how the RBM preferentially move to higher probability states and stochastically moves through the state space.  Each unit has a sigmoidal activation with activation probability $p(v_i = 1 | h) = \sigma(w_i^Th+b_i)$ and $\sigma(x) = (1 + e^{-x})^{-1}$. 

\subsection{Merging RBMs}
\label{sec:merge}

The difficulty of training large RBMs means that new innovations are necessary at the training step. In this work, we propose merging smaller models, that are already trained, to form an initial condition for larger models as a way of improving the training of large RBMs. This methodology is inspired from the digital logic where larger functions can be constructed by combining small functional blocks \cite{Camsari2017StochasticLogic}.  Notably, all NP-hard problems can be formulated through the Boolean Satisfiability problem. Therefore, constructing the RBM in the aforementioned way provides a natural approach to solving hard optimization problems. An example of this is shown in Figure \ref{fig:alg} F) where we solve a toy example of a 3SAT Boolean Satisfiability problem. This shows that this method has the representational power to solve a wide variety of NP-Hard problems. 

Merging is performed by combining RBMs along their common visible neuron connection. We show the mechanics of the merging process in Figure \ref{fig:alg} C) and D) where we combine digital logic gates together in this manner. Merging across the visible neurons like this retains the bi-partite, \emph{product of experts} nature of the RBM while giving the expected distribution if we were combining gates to perform logical synthesis. Using this as inspiration, we construct adders and multipliers as shown in Figure \ref{fig:alg} E) to combine trained n-bit adder and multiplier units into 2n-bit multiplier units. Detailed information of this merging protocol has been provided in the Methods section. 

After smaller models have been trained and merged, we retrain the larger models to fine-tune them. As shown in Methods, Equation \ref{eq:KL}, the models are good approximations for the correct distribution we are interested in, and provide a good initial conditions for training of the final model. Importantly, merging in this way retains the bi-directional nature of the network. The same RBM can be queried to solve what the output is for a given input (the ``forward'' direction), and queried what outputs caused a given input (the ``reverse'' direction). In figure \ref{fig:fact} A), B), and C) we see the consequence of the bi-directional nature where the same model can perform multiplication, division and factorization, as it learns the full joint distribution over variables. In these tasks, the model must get the exact solution as the mode of the sampled distribution. Performance is reported as $p_{correct}$ which is the fraction of instances which report the correct mode for 300 randomly generated instances after the given number of samples. We sample for the given number of samples and check whether the mode of the sampled distribution corresponds to the correct answer to the problem of interest. We additionally show in those figures that by merging and retraining we get significantly better performance than training alone or merging alone.

\section*{FPGA Acceleration}
\label{sec:FPGA}
The massive parallelism present in the RBM algorithm makes it especially efficient on the FPGA. The RBM algorithm also doesn't contain any branches or explicit memory accesses while sampling, removing expensive branch misprediction cycles and DRAM fetch cycles. Furthermore, unlike other deep neural network accelerators, this algorithm is not memory bandwidth limited for any of its operation \cite{Jouppi2017In-DatacenterUnit} as can be seen by the FPGA utilization table (Table \ref{tab:utilization} in Supplementary Section), further increasing the algorithmic performance on hardware. The bipartite nature means that many neurons can be sampled in parallel on the FPGA, allowing us to perform each neuron activation probability in parallel. There has been much work on accelerating RBM training through FPGA implementations \cite{Ly2009AMachines, Kim2009AImplementation, Kim2010AMachines}, but by focusing on inference only, we reduce the necessary hardware requirements to the essential components, fully unlocking the inherent parallelism in the network architecture.

\subsection{Model Quantization} 
\begin{figure*}
\begin{centering}
\includegraphics[width=\linewidth]{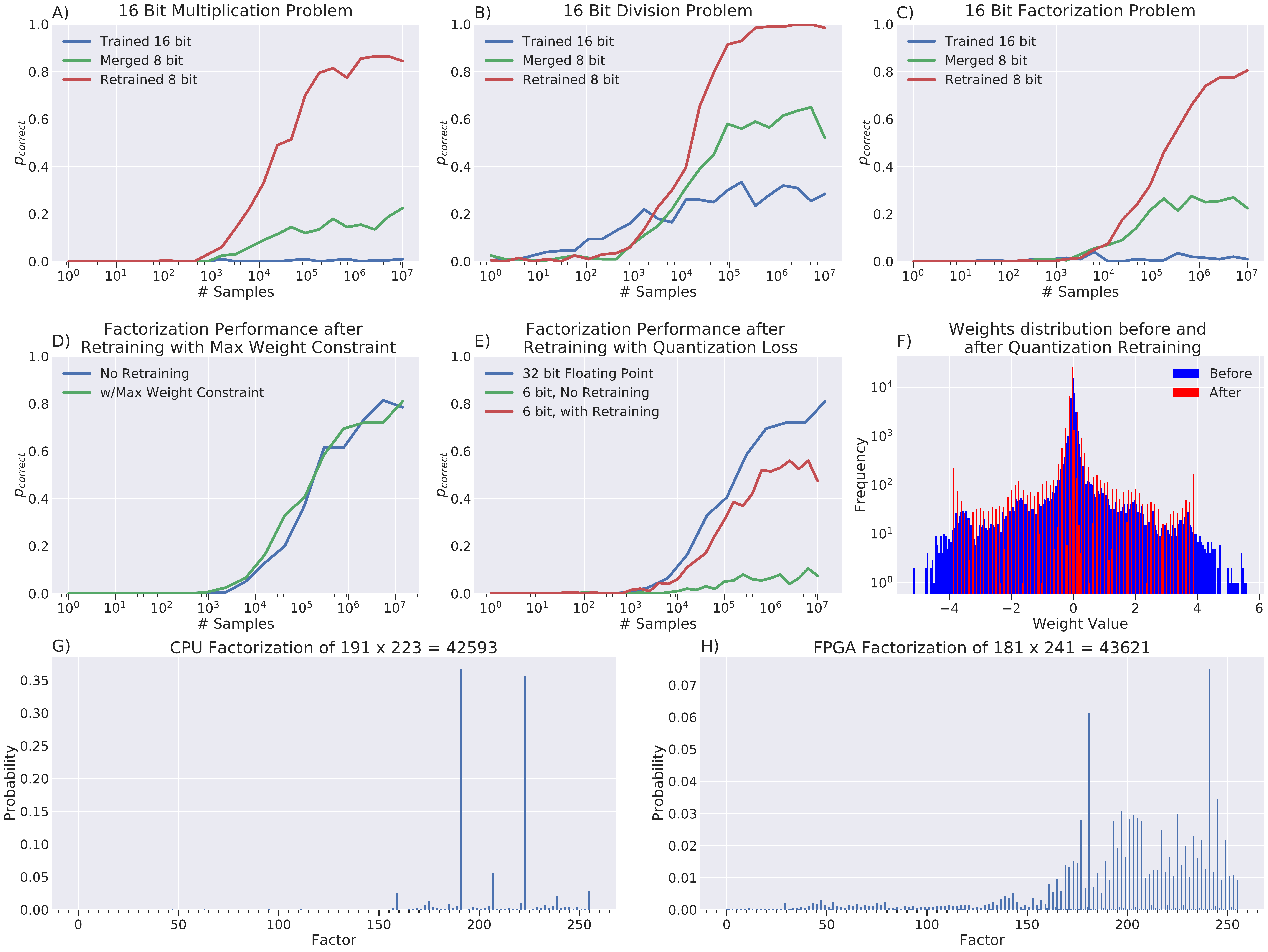}
\par\end{centering}
\caption{\label{fig:fact} {\bf Performance on 16 Bit Multiplication, Division and Factorization} \\ 
{\bf (A), (B), (C)\/} Showing the performance on multiplication, division, and factorization performed by directly training a 16 bit network (trained 16 bit), merging two 8 bit networks (merged 8 bit), or merging two 8 bit networks and retraining (retrained 8 bit). 
{\bf (D)\/} Showing  the effect of retraining with a maximum weight constraint. Here we see no performance degradation due to retraining the module by adding this extra constraint.
{\bf (E)\/} Retraining the network with added $L_1$ quantization loss. By retraining for 6 bit quantization, we see a large increase in performance compared to naive quantization.
{\bf (F)\/} A histogram of the RBM weights before and after retraining for quantization. We see the network is strongly clustered around the 6 bit values.
{\bf (G)\/} An example of a factorized distribution after $10^7$ samples showing factorization of a 16 bit number into its two prime factors. The sampled distribution shows clear peaks at the two correct answers to the factorization problem. 
{\bf (H)\/} An example probability mass function showing what the sampling procedure would return after running on the FPGA for $10^7$ samples. Although this shows that there is greater error in the incorrect factors as compared to part (G), there are still two clear peaks in the distribution indicating the correct factors. 
}
\end{figure*}

In addition to taking advantage of the inherent parallelism, to fit larger models on the FPGA we have performed model quantization to be able to lower precision during the inference. There has been much work on model quantization for deep neural networks, however most of them focus on Convolutional Neural Networks and Multi-Layer Perceptrons \cite{Han2016DeepCoding, Ullrich2019SoftCompression, Chen2015CompressingTrick}, which cannot be directly used for RBMs. We developed a scheme for model quantization, explicitly for RBMs, which is accomplished via 2 additional training steps. The first is adding a constraint to the maximum value of the weight and retraining with this constraint. This makes sure that the weight magnitude cannot overflow the fixed point representation that we are trying to accomplish. As demonstrated in Figure \ref{fig:fact} D), this does not change the overall accuracy of the model, as the retraining causes more weights of smaller magnitude to compensate for a single weight of large magnitude. The second retraining step adds an extra quantization loss term to the training step (see Eqn. \ref{eq:quant_decay} in Methods). In Figure \ref{fig:fact} E) we show that we can quantize from 32 bit floating point to 6 bit fixed point without large loss in error. Figure \ref{fig:fact} F) shows the final weights distribution after these two quantization training steps, demonstrating how the retraining preferentially pushes weights towards their post-quantization values. The exact training steps are detailed in the Methods section. In Figure \ref{fig:fact}  H) we see how this quantization effects the performance of a factorization task. This increase in error is accompanied by massive increases in speed and power efficiency. 

\subsection{Matrix Multiplication with Mask}
With the use of a fixed point representation of the model parameters and our restriction of binary node values, the matrix multiplication step of the algorithm is greatly simplified. Instead of instantiating many binary multiplication circuits, we can perform the computation by passing the weight values through a binary mask, or a series of 2-to-1 muxes, created by the node values. We combine this with the use of fixed point rather than floating point calculations, and we see a large decrease in area cost and FPGA resource usage. The estimated area cost of 32 bit floating point multiplication is 27x that of an 8 bit multiplication, and an 8 bit multiplication is 8x costlier than an 8 bit adder. This shows that the benefits of getting rid of the multiplications is very large, allowing more of the calculation to happen on the FPGA in one cycle \cite{Dally2015High-PerformanceLearning}. 

\subsection{Sigmoid Approximation}
Exact calculation of the sigmoidal activation function $f(x) = \frac{1}{1 + e^{-x}}$ is computationally expensive. To accomplish direct calculation, at least 3 extra hardware instructions are needed, exponentiation, addition, and division, which all incur a large hardware cost both in terms of latency and area. Instead, binary sigmoid values are precomputed and enumerated in a look up table (LUT) for use in the FPGA.  This implementation allows for fast evaluation of the activation function without expensive hardware resources. After matrix multiplication and bias addition, the computed value is passed through the LUT based activation function to approximate the sigmoid. 

\subsection{Pseudo-Random Number Generation}
We have found that the quality of the random numbers is not important to the algorithm working effectively. The only necessary condition was that the random numbers were not correlated between individual neurons. To accomplish this we use a 32 bit length Linear Feedback Shift Register (LFSR) pseudo random number generator to create high quality pseudo-random numbers. The total cost of these LFSR based random number generators amounts to just 5\% of the design flip flop usage, and 2\% of the lookup table usage making them relatively cheap in comparison to other higher quality random number generators. Each neuron has its own LFSR and is seeded with a different value to minimize the possibility of correlation. Longer sampling runs ($>10^8$ samples) would cause the neuron LFSRs to start looping, and correlate samples at the start of the run with later samples but this can be simply mitigated by increasing the length of the LSFR. 

\section*{Results and Discussion}

\subsection{Trained Model Performance}

By merging together small RBMs using the principles of logic synthesis, many complex and NP-Hard problems can be solved using the bi-directional nature of RBMs. In Figure \ref{fig:alg} D) we see the mechanics of this procedure, while Figure \ref{fig:alg} E) shows how it can apply to integer factorization and Figure \ref{fig:alg} F) shows how it can apply to 3SAT and boolean satisfiability. We note that the logic gates formed in part \ref{fig:alg} C) can be operated in reverse to find solutions to the given satisfiability problem with 3SAT. This is the canonical example of an NP-Hard problem, which maps directly to all other NP-Hard problems, showing that these networks have the representational power to solve a variety of such problems. \cite{Cook1971TheProcedures, Karp1972ReducibilityProblems}. These problems are at the heart of many computationallly difficult problems. We additionally show how this type of training can outperform directly trained models. Figure \ref{fig:fact} A), B) and C) show that merging smaller models and retraining them drastically outperforms directly training the model, with errors close to 3-10x less for problems of interest. In Figures \ref{fig:fact} G) we show how these models can factor a semi-prime number into its two factors with high accuracy. The factorization of a semi-prime number into its two co-primes is at the heart of the RSA cryptography algorithm and is the basis of most of modern encryption systems. 

To take advantage of the parallel resources of the FPGA and unlock hardware performance, we have shown generalizable methods for quantization that can be applied to all RBM problem instances. This is shown in figure \ref{fig:fact} D), E) and F) where we show that adding a max weight constraint followed by quantization loss while training can lead to better performance once quantized on the FPGA. By showing significant model performance down to 6 bit integer representations, we demonstrate that this approach allows for hardware efficient model representations for RBMs.  

\subsection{Scaling} 
\begin{figure*}
\begin{centering}
\includegraphics[width=0.95\linewidth]{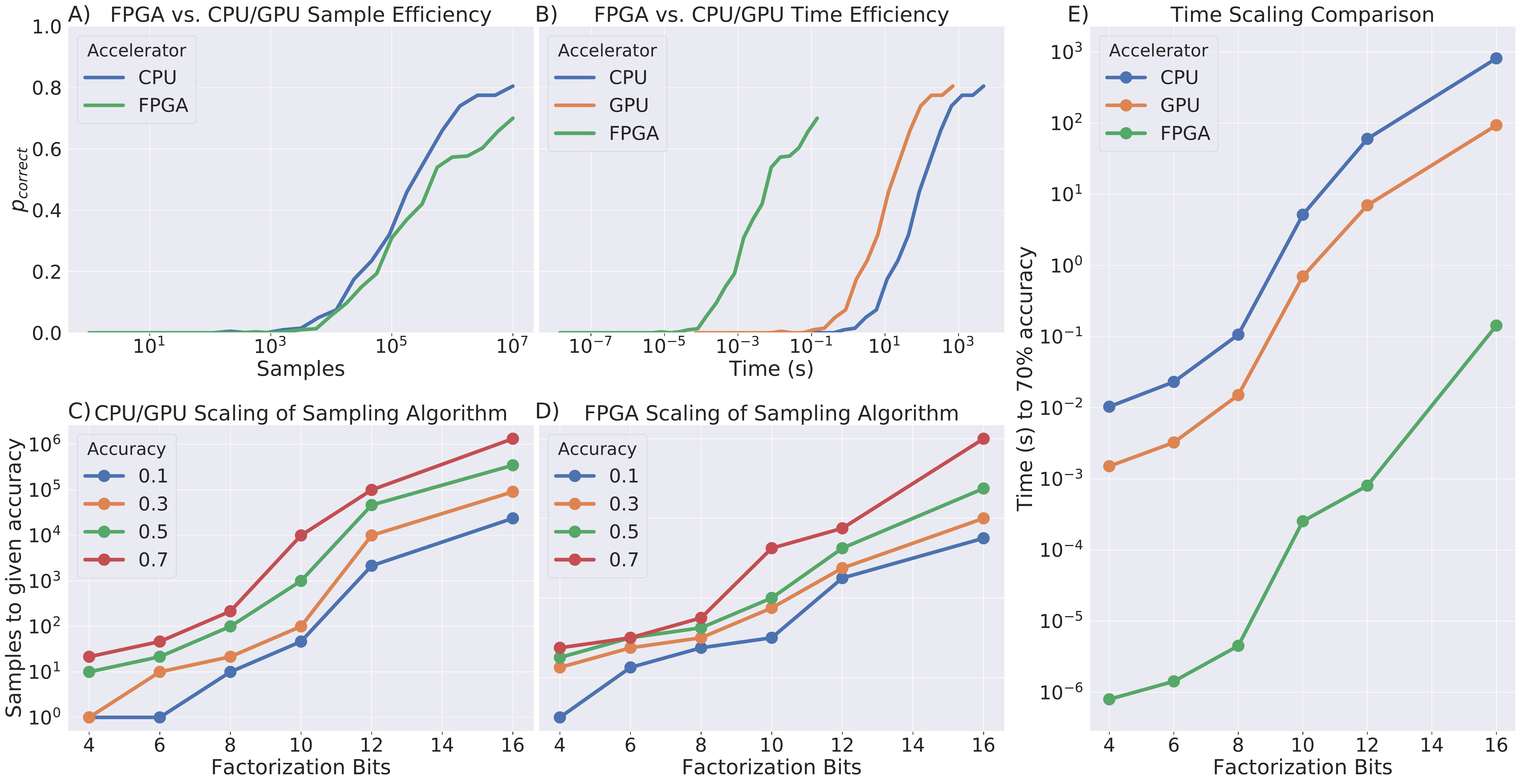}
\par\end{centering}
\caption{\label{fig:fpga_perf} \textbf{Performance of the FPGA implementation vs the CPU implementation on factorization} \\
The sampling algorithm scales approximately exponentially with the bit size (and approximately linearly with the phase space). We see a $10^4$ speed improvement across all model sizes compared to the CPU algorithm and $10^3$ speed improvement compared to the GPU algorithm. 
{\bf (A)\/} The sample efficiency of the FPGA implementation is similar to the CPU implementation, even after quantizing to 8 bit weights and biases and using the various approximation schemes detailed. 
{\bf (B)\/} When the time taken to reach a solution is scaled for the FPGA vs. the CPU and GPU, the FPGA outperforms both by orders of magnitude., 
{\bf (C)\/} The scaling of the algorithm when measured at various accuracy levels on the CPU. The RBM for each bit number is run until it hits the given accuracy on a set of random factorization problems. 
{\bf (D)\/} Scaling of the sampling algorithm when run on the FPGA. The difference in sample number from part (C) is due to approximations necessary to efficiently port the model onto the FPGA. 
{\bf (E)\/} Time scaling of the factorization problem measured at the 70\% accuracy level. We see that the FPGA performs 4 orders of magnitude faster compared to the CPU and 3 orders of magnitude compared to the GPU across all bit counts for the outlined sampling algorithm.
}
\end{figure*}

We have demonstrated scaling of the factorization algorithm up to 16 bit numbers. Markov Chain Monte Carlo based sampling methods for optimization problems fall into the class of "Stochastic Local Search" and are expected to have exponential scaling with problem size \cite{Hoos2005StochasticSearch}. This exponential scaling dependence is shown in Figures \ref{fig:fpga_perf} C) and D). Although this is the case, we see a $10^4$ constant factor speed  increase when the algorithm is implemented on the FPGA in Figure \ref{fig:fpga_perf} B) in 16 bit factorization and Figure \ref{fig:fpga_perf} E) across all problem instances. This massive speed increase across the whole spectrum of bit sizes has real world consequences, as it implies that other algorithms mapped onto this general framework can become very efficient in finding ground state solutions which would otherwise be difficult to obtain.

The FPGA implementation of our sampling algorithm has shown a $10^4$ speed increase compared to a dual CPU system, and a $10^3$ speed increase compared to a GPU. This comes with a 32x power decrease compared to the CPU and 6x power decrease compared to the GPU. This is compared to a dual CPU machine running the highly optimized industry standard PyTorch machine learning framework, and a hand optimized GPU algorithm using the CUDA and cuBLAS libraries. We note that the performance improvement of the GPU algorithm compared to the CPU is minimal due to the thread synchronization, limited cache sizes and relatively small RBM size presented here.\cite{Ly2008NeuralMachines} Although our implementation takes up much of the resources of the FPGA, there are many possible areas where our design could be modified to scale for performance. Our goal was not to create the most optimized hardware design, but to demonstrate that parallel hardware running our very hardware friendly algorithm had the potential for drastic improvement.  With focused effort on the improvement of the hardware architecture \cite{Li2015AnStreams,Ly2009AMachinesb, Lo2011BuildingMPI} the speed and performance improvement is expected to get much larger. 

Although we demonstrate sampling speed increase for inference, using the same FPGA accelerated sampling algorithm can also work for decreasing training time. The major bottleneck in the training step is creating a series of uncorrelated samples, which takes a large number of samples for a highly correlated sampler. Using FPGA acceleration of the sampling algorithm could give a lower variance estimate of model probabilities in a much faster and energy efficient manner, than those provided by a CPU or GPU.

\subsection{Time Domain Analysis}

Markov Chain Samplers (and specifically the Gibbs Sampler we are using) have been proven to converge in a geometric rate with the number of samples. In addition, the distance in total variation between the sampled distribution and the model distribution strictly decreases with each Gibbs sampling step \cite{Bremaud1999GibbsSimulation}. This implies that the quality of our solution should increase as the sampled distribution approaches equilibrium. This is especially useful for optimization problems where run time can be traded directly for better solutions. 

\begin{figure*}
\begin{centering}
\includegraphics[width=0.88\linewidth]{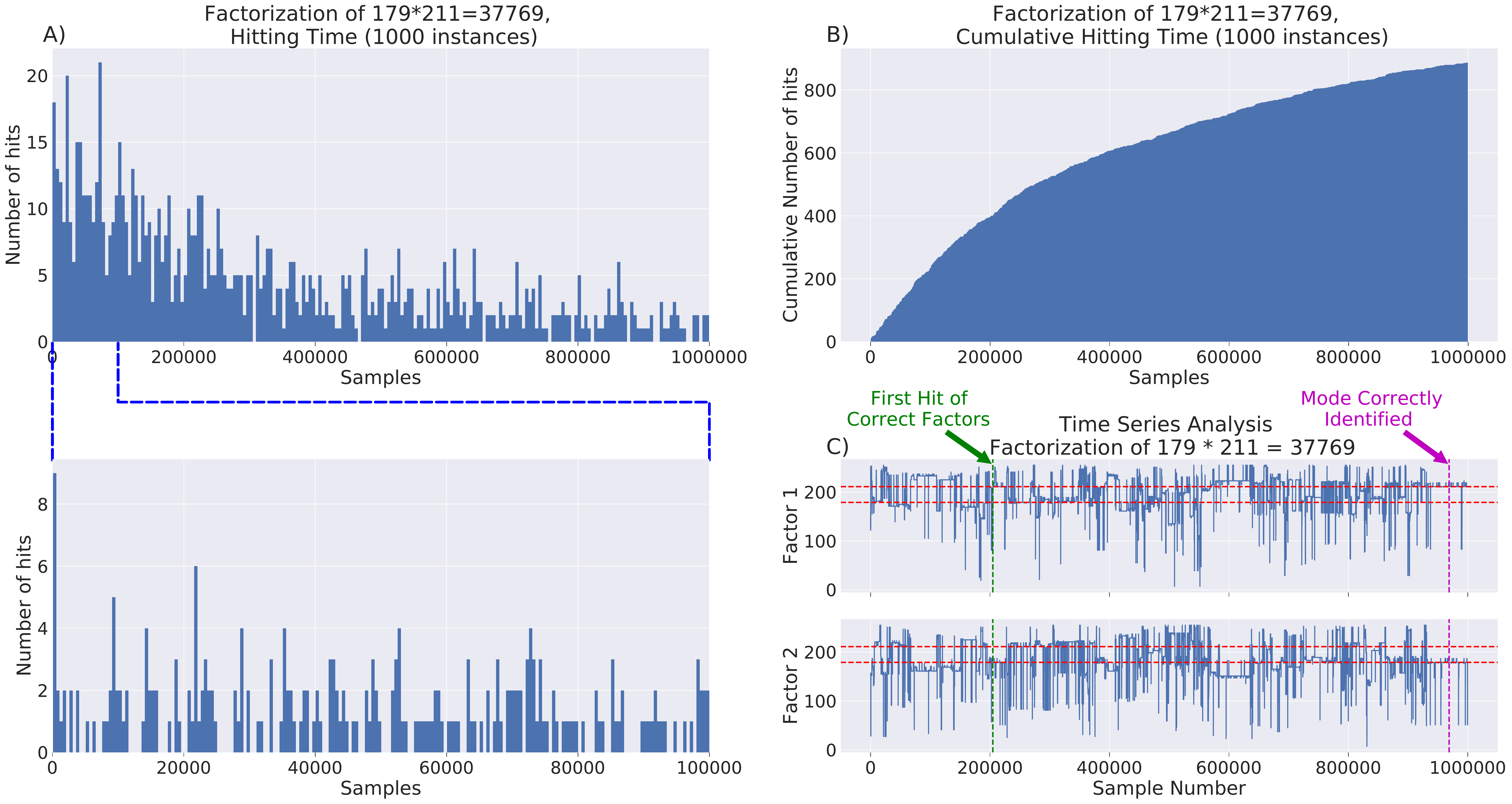}
\par\end{centering}
\caption{\label{fig:hit}{\bf Time Domain Analysis of 16 Bit Factorization Algorithm}  \\
{\bf (A)\/} Hitting time histogram for the factorization of a 16 bit number. The experiment is repeated 1000 times, and aggregate results are shown for this. This shows that most hits happen fairly early, with a long tail to the distribution. On the bottom panel, when zooming into the hits that occur in the first 100000 samples, we see that there are a large number of hits that occur in the first bin. 
{\bf (B)\/} Histogram showing the cumulative distribution for the hitting time on the 16 bit factorization task. The smoothness of this distribution can allow us to fit an empirical model for the samplers and effectively parallelize by starting many samplers. 
{\bf (C)\/} A trace plot showing the time domain behavior of the sampler. The sampler stochastically explores the state space, but preferentially remains in the minimal energy state. This means that the first hit of the correct factors occurs much quicker than the minimal energy/maximal probability state is correctly identified.  }
\end{figure*}

The time domain sampler also shows a large difference between time taken to model the full distribution (the mixing time) and the time taken to sample the correct factors once (the hitting time)(see Fig. \ref{fig:hit} A and B in comparison to Figure \ref{fig:fpga_perf} C and D). A demonstrative example of this phenomenon is shown directly in Figure \ref{fig:hit} C), with the difference in times being close to 4x. By adding in a sample verification methodology, the time taken to identify correct factors can be drastically decreased. For NP-Hard problems, the solution can be verified in polynomial time but finding a given solution is done within exponential time, a feature which is also present in this factorization problem. This means the relative overhead of checking of factors is relatively low, and can be done on a regular basis to reduce the sampling time significantly. This method of adding heuristics to the sampler is present in many different problem types, and must be done differently for each type of problem. Here we demonstrate both that the sampler approaches the correct distribution, and that effective heuristics using the hitting time are available to reduce the time to solution.

\section*{Conclusions}
\label{sec:conclusion}
\begin{figure*}

\begin{centering}
\includegraphics[width=\linewidth]{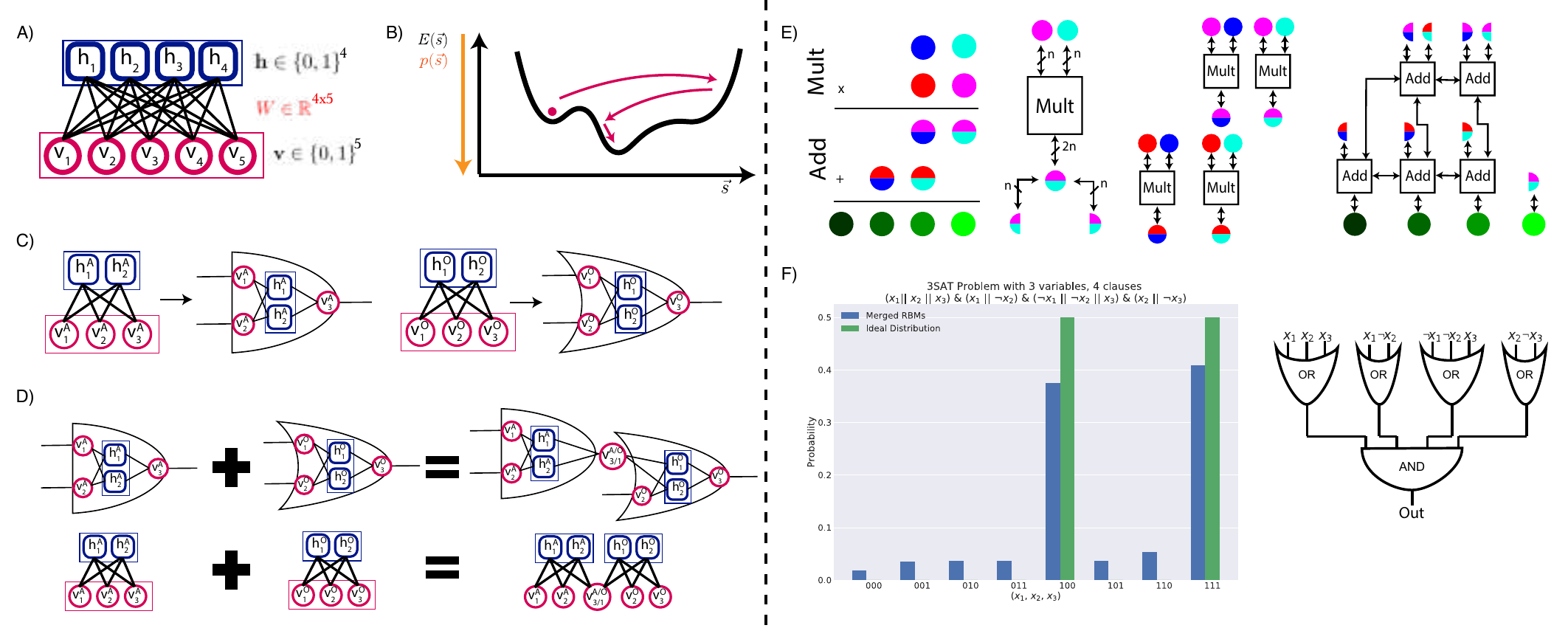}
\par\end{centering}
\caption{\label{fig:alg}. \textbf{Demonstration of RBM structure and sampling algorithm} \\ 
{\bf (A)\/} Structure of the RBM neural network. The Restricted Boltzmann Machine is a binary neural network structured in a bipartite graph structure.
{\bf (B)\/} The RBM maps out the non-convex state space of a probability distribution. Low energy states map to high probability states which the network identifies through a Markov Chain Monte Carlo (MCMC) algorithm. 
{\bf (C)\/ } A graphical mapping of RBMs to gate level digital circuits.  The visible nodes correspond to the inputs and outputs of the logic gate, and the hidden nodes are the internal representation of the logic gate. 
{\bf (D)\/} Graphical Demonstration of the merging procedure, showing how two RBMs which represent an AND gate and an OR gate can be merged together to form a connection.
{\bf (E)\/} We can create arbitrary adders and multipliers by merging together smaller units to create the logical equivalent of larger units. The leftmost image shows how we create a 2n bit multiplier using n bit multiplications and n bit additions. The color coding shows how the partial products are broken apart amongst the adders and multipliers. To the right of that we show how we perform 4, $n$-bit input 2$n$-bit output multiplications, and then accumulate the result.
{\bf (F)\/} Using this strategy of merging logical units to solve a simple 3SAT, Combinatorial Optimization problem. 
}

\end{figure*}

Our choice of looking at an integer factorization problem stems from two motivations: (i) as we are exploring a new training methodology, the factorization problem gives access to large amount of training data without worrying about quality and (ii) at the same time integer factorization is also an example of a hard problem. It should be noted, nonetheless, that many combinatorial optimization problems can be broken down into associated sub-problems, and solved using a greedy approach (i.e. using the nearest neighbor approach in the Travelling Salesman Problem, or multiplying single digits in a larger multiplication, or evaluating one Boolean logic statement in a Boolean satisfiability problem as shown in Figure \ref{fig:alg}). Using a greedy approach (such as the one used in the travelling salesman problem with nearest neighbors) can produce non-optimal results, and evaluating all permutations of possible solutions to find the most optimal one can be computationally intractable in a large problem space. By combining these sub-problems using the method proposed here, we bypass the problems associated with those two approaches.  We encode possible solutions as a probability indicating its local optimality, and combine these sub problems by merging the visible units of their RBMs. This combination mechanism multiplies the probabilities such that the solution with global optimality is encoded as the mode of the distribution modeled by the larger RBM. In addition, as the phase space of the problem space is $2^{v}$ where $v$ is the number of visible units, we can also encode a large problem space using minimal units and decrease the hardware cost of the approach. As the Boolean satisfiability problem (shown by the 3SAT example) can be mapped to solving a variety of NP-Hard problems \cite{Cook1971TheProcedures, Karp1972ReducibilityProblems}, it is expected that variants of this RBM framework can be applied to a large variety of graph based optimization tasks such as MAX-CUT, the Travelling Salesman Problem, Quadratic Assignment problem and any problem that can be mapped to a series of binary variables. 

 Much work has been done to develop accelerators for the Ising Model problem, such as specialized ASICs \cite{Yamaoka2016AAnnealing, Boyd2018SiliconNews, Schneider1993AnalogCircuits}, FPGA designs \cite{Belletti2009Janus:Computing, Ko2019Flexgibbs:Graphs}, Memristor based accelerators \cite{Bojnordi2016MemristiveLearning, Wan2020AModels}, Quantum Mechanical Accelerators based on quantum adiabatic processes \cite{Dridi2017PrimeGeometry}, Optical Parametric Oscillators \cite{Wang2013CoherentOscillators, McMahon2016AConnections}, Magnetic Tunnel Junction \cite{Camsari2017StochasticLogic, Camsari2017ImplementingMTJ, Borders2019IntegerJunctions} and many others. Our work is distinct from these existing results in that, we showed a new way of  constructing large models from smaller building blocks. We further demonstrated that its implementation on an FPGA, which exploits the intrinsically parallel architecture and sparsity, can lead to orders of magnitude improvement in the speed and power for a sufficiently large problem (2$^{32}$ phase space). Importantly, it should be noted that these improvements in performance cannot be decoupled from the algorithmic advances that we have proposed.  At the same time, our approach could also benefit from the emerging hardware proposed in the aforementioned reports for further improvement in speed and energy. Further scaling can also be accomplished by moving to Deep Boltzmann machines \cite{Salakhutdinov2009DeepMachines, Salakhutdinov2010EfficientMachines}, which may be easier to train.  While we demonstrated an inference problem, the underlying method is equally applicable and expected to accelerate training problems \cite{Dally2015High-PerformanceLearning, Lo2011BuildingMPI, Savich2011ResourceMNIST}. Substantial acceleration of generative models such as the RBM could lead to unsupervised, life-long, learning machines.


\clearpage

\clearpage

\clearpage

\clearpage


\clearpage
\newpage
\begin{methods}

\begin{subsection}{Model Training}
In this paper we trained the RBMs by contrastive divergence as described by \cite{Hinton2002TrainingDivergence}. Each of the models in this paper were validated by checking their performance on the problem they were trying to solve (i.e addition and subtraction for an adder, multiplication and factorization for  a multiplier).  This method was also used to assess model complexity (i.e number of hidden units) and evaluate learning parameters (learning rate, batch size, etc.). The final results for RBM sizes and approximate training times are shown in Table \ref{tab:models}. Models were trained using the PyTorch library in Python, which is one of the standard libaries used in Machine Learning research and production code. 

Training was conducted on a computer with 2 Intel Xeon E5-2620 processors, and 2 Nvidia Titan V GPUs. Each RBM was trained until the test error stopped decreasing, and the model had converged to a best solution. Training rate was varied across models and tuned as a hyperparameter. The training set was copies of all possible multiplication or addition problems, encompassing the full state space. In the case that the state space was too large to train on all of it (such as the 16 bit and 32 bit adders), a random sample of the training set was used, and a new random sample was reinitialized every epoch of training.   

The training time tends to increase with the number of bits in the adder or  multiplier due to both the size of the data set (which increases exponentially with the number of bits) and the number of hidden and visible units (which both increase approximately linearly). 

At the 16 bit adder level, the size of the data set was so large that the entire data set could not be used for training ($\approx 8$ billion data points) and a randomized sample of the set had to be taken. As generalization is not perfect, we can attribute the decrease in their performance to this fact. For the 32 bit adder, this problem was exacerbated, and the 32 bit adder was outperformed by most units even after training for a full week. 

For multipliers, the 8 bit multiplier has a tractable amount of data, but a good joint density model could not be formed even after a large training time. We believe this is due to an inherent difficult in the multiplication problem that is not present in the addition problem. The multiplication model has a higher number of hidden units as there is not as distinct of a correlation between higher level bits in the 8 bit multiplication problem as there is in the addition problem, first level correlations (as an RBM with 1 layer of hidden units would find) are more difficult to find. We believe that using deep boltzmann machines might help fix the problem of training in large multipliers. 
\end{subsection}

\begin{subsection} {Merging RBMs}
Given two RBMs we wish to merge along a common connection (see Figure \ref{fig:alg}) with the following parameters: $W_A \in \reals^{n\times r}$ and $W_B \in \reals^{m\times s}$, visible biases $b_A \in \reals^{n}$ and $b_B \in \reals^{m}$, and hidden biases $a_A \in \reals^{r}$ and $a_B \in \reals^{s}$. The energies and probabilities of these are as follows:
\begin{equation} \label{eq:energies}
\begin{split}
E_A(v, h) &= -v^TW_Ah - a_A^Th - b_A^Tv; \;\; p_A(v, h) = \frac{1}{Z_A} e^{-E_A(v, h)} \\
E_B(v, h) &= -v^TW_Bh - a_B^Th - b_B^Tv; \;\; p_B(v, h) = \frac{1}{Z_B} e^{-E_B(v, h)}  
\end{split}
\end{equation}

We can write the weight matrices as a series of row vectors corresponding to one visible unit's connections to a set of hidden units. 

 \begin{equation} \label{eq:weights}
\begin{split}
W_A = \begin{bmatrix}  \rule[.5ex]{1em}{0.4pt}   w^A_1  \rule[.5ex]{1em}{0.4pt} \\ 
 \vdots \\ 
 \rule[.5ex]{1em}{0.4pt}   w^A_n  \rule[.5ex]{1em}{0.4pt} \end{bmatrix}, \;\;  
   W_B = \begin{bmatrix}  \rule[.5ex]{1em}{0.4pt}   w^B_1  \rule[.5ex]{1em}{0.4pt} \\ 
 \vdots \\ 
 \rule[.5ex]{1em}{0.4pt}   w^B_m  \rule[.5ex]{1em}{0.4pt} \end{bmatrix}, \;\; \\ 
 \end{split}
\end{equation}
   
 With this definition, the merge operation is shown below. If unit $k$ of RBM $A$ is merged with unit $l$ of RBM $B$ the associated weight matrix $W_{A+B}  \in \reals^{(n+m-1)\times (r+s)}$ , visible bias $v_{A+B} \in \reals^{n+m - 1}$ and hidden bias $h_{A+B} \in \reals^{r+s}$ dictate the probabilities and energies for the merged RBM. Merging multiple units between these two RBMs corresponds to moving multiple row vectors from $W_B$ to $W_A$, which creates the associated decrease in dimensionality of $W_{A+B}$ and $b_{A+B}$ (where $W_{A+B}  \in \reals^{(n+m-d)\times (r+s)}$ and $v_{A+B} \in \reals^{n+m - d}$ where $d$ is the number of merged units.

 \begin{equation} \label{eq:mergedweights}
W_{A+B} = 
 \left[\begin{array}{@{}c | c@{}}
 \mbox{\normalfont $W_A$}
  & \begin{matrix}
  \mbox{\normalfont 0} \\ 
  \rule[.5ex]{1em}{0.4pt}   w^B_l  \rule[.5ex]{1em}{0.4pt} \\ 
  \mbox{\normalfont 0} \\ 
  \end{matrix} \\
 \hline
  \bigzero &
 \mbox{\normalfont $W_{B\setminus l}$}
\end{array}\right] = 
 \left[\begin{array}{@{}c | c@{}}
 \begin{matrix}
 \rule[.5ex]{1em}{0.4pt}   w^A_1  \rule[.5ex]{1em}{0.4pt} \\
  \vdots \\ 
  \rule[.5ex]{1em}{0.4pt}   w^A_k  \rule[.5ex]{1em}{0.4pt} \\ 
  \vdots \\ 
  \rule[.5ex]{1em}{0.4pt}   w^A_n  \rule[.5ex]{1em}{0.4pt} \\
  \end{matrix}
  & \begin{matrix}
 \bigzero \\ \\
  \rule[.5ex]{1em}{0.4pt}   w^B_l  \rule[.5ex]{1em}{0.4pt} \\ \\
  \bigzero \\ 
  \end{matrix} \\
 \hline
  \bigzero &
 \begin{matrix}
 \rule[.5ex]{1em}{0.4pt}   w^B_1  \rule[.5ex]{1em}{0.4pt} \\
  \vdots \\ 
  \rule[.5ex]{1em}{0.4pt}   w^B_{l-1}  \rule[.5ex]{1em}{0.4pt} \\ 
  \rule[.5ex]{1em}{0.4pt}   w^B_{l+1}  \rule[.5ex]{1em}{0.4pt} \\ 
  \vdots \\
   \rule[.5ex]{1em}{0.4pt}   w^B_m  \rule[.5ex]{1em}{0.4pt} \\
  \end{matrix}
\end{array}\right] \\
\end{equation}
\begin{equation} \label{eq:mergedbiases}
 b_{A+B} = \begin{bmatrix} b^A_1 \\ \vdots \\ b^A_k + b^B_l \\  \vdots  \\ b^A_n \\  b^B_1 \\ \vdots \\ b^B_{l-1} \\ b^B_{l+1} \\ \vdots \\ b^B_{m} \\ \end{bmatrix}  \; \; 
a_{A+B} = \begin{bmatrix} a^A_1 \\ \vdots  \\ a^A_r \\  a^B_1 \\ \vdots \\  \\ a^B_{s} \\ \end{bmatrix}
\end{equation}

Below, we show how this relates to the original energies and probabilities. The vectors $v$ and $h$ correspond to the visible vector put into the combined RBM, while $v_A$, $v_B$, $h_A$ and $h_B$ correspond to the equivalent state vectors that would be inputted into the single RBMs. Using these equations, we can see that the combined RBM energy factorizes into a sum of the original RBM energies and the probability is the product of the original probabilities. 
 \begin{align} \label{eq:mergedvis}
v & = \begin{bmatrix} v_1 \\ \vdots \\ v_{n+m-1}\end{bmatrix} \;\; 
h = \begin{bmatrix} h_1 \\ \vdots \\ h_{r+s} \end{bmatrix} \;\; \\
v_A & = \begin{bmatrix} v_1 \\ \vdots \\ v_l \\ \vdots \\ v_n\end{bmatrix}, v_B = \begin{bmatrix} v_{n+1} \\ \vdots \\ v_{n+l-1} \\ v_{l} \\ v_{n+l} \\ \vdots \\ v_{n+m - 1}\end{bmatrix}, \\ 
h_A & = \begin{bmatrix} h_1 \\ \vdots \\ h_r \end{bmatrix} h_B = \begin{bmatrix} h_{r+1} \\ \vdots \\ h_{r+s} \end{bmatrix} 
\end{align}
\begin{equation} \label{eq:mergedEnergy}
    E_{A+B}(v, h) = E_A(v_A, h_A) + E_B(v_B, h_B), 
\end{equation}
 \begin{equation} \label{eq:mergedprobs}
  p_{A + B}(v, h) = \frac{1}{Z_{A+B}} e^{-E_{A+B}(v, h)} \propto p_A(v_A, h_A)p_B(v_B, h_B) 
 \end{equation}
Because of the probabilities approximately multiplying (Eq. \ref{eq:prob_mult}), we can also say that if each of the distributions differed from the ``ideal'' distribution (denoted here by $q$), then we can expect the error (as measured by the KL divergence eqn. \ref{eq:KL}) to increase approximately linearly with the number of distributions summed together. As contrastive divergence learning approximately follows the gradient of the KL divergence, the merged model represents a good initial condition for training of the larger model \cite{Carreira-Perpinan2005OnLearning}. This means that only small corrections in CD training are needed on the merged model to create a good trained model for the larger network. Training the merged model is possible as intermediate nodes are represented as extra visible units, and can be calculated based on the input and outputs of the dataset. The data for the merged models can be calculated as if we were propagating values through a digital circuit and keeping track of intermediate values, which become the data for the merged model to be trained on. 
\begin{align}
 \label{eq:prob_mult}
  \quad p =  p_A(v_A, h_A)p_B(v_B, h_B),  \;\;  q & =q_A(v_A, h_A)q_B(v_B, h_B) \\
 \label{eq:KL}
       \KL(p \| q)  \approx  \KL(p_A \| q_A) & + \KL(p_B \| q_B);
 \end{align}
\end{subsection}

\begin{subsection}{Retraining for Quantization}

To train for quantization, the loss function that is optimized for ($L(W, d)$) is modified so that in addition to having the regular Contrastive Divergence loss between the weights $W$ and the data $d$ denoted by $CD(W, d)$, we have a loss term that pushes weights to be closer to their quantized value.  The hyperparameter $\lambda$ is slowly increased during training to force the weights progressively closer to their quantized value. This method allows the contrastive divergence term to fix the errors created by quantizing the weights slowly while training.  Although taking the exact gradient of this loss term is not possible (as $Q(W)$ is not a smooth function of $W$, see  Eqn. \ref{eq:quant_decay2}), we find that by assuming the quantization gradient $\frac{\partial Q(W)}{\partial W} \approx 0$, we obtain a sufficiently good performance. 
\begin{align}
\label{eq:quant_decay}
    L(W, d) & = \epsilon CD(W, d) - \lambda ||W - Q(W)||_1 \\
\label{eq:quant_decay2}
    \frac{\partial L(W, d)}{\partial W} & = \epsilon \frac{\partial CD(W, d)}{\partial W} - \lambda sign(W - Q(W)) (1 - \frac{\partial Q(W)}{\partial W})
\end{align}

\end{subsection}

\begin{subsection}{FPGA Programming}
At the heart of the FPGA is the RBM computing core which performs the Gibbs sampling algorithm. All programming was done using the Xilinx Vivado suite on the on the Xilinx Virtex UltraScale+ XCVU9P-L2FLGA2104. The core was designed to output the most samples for the RBM sizes we had.  The weights and biases are stored in on-chip SRAM to decrease access time. The values are broadcast to the node update modules each cycle, which performs the necessary operations for the sampling and take up the bulk of the computation resources. There are no pipeline or data hazards, removing the need for any complex timing schemes.  Thus, if we instantiate a node update module for every node register, there is a new sample taken from the visible node registers every clock cycle, taking full advantage of the RBM's parallelism.

Each node update module contains the logic to perform a matrix-row multiplications, a sigmoid function, and a comparison with a random number. The matrix-row multiplication is performed via a binary mask and an adder module that accumulates the surviving weight values. Once each weight value is masked appropriately, they are passed into an adder tree which accumulates the results of each multiplication. The accumulators take the majority of LUT resources on the FPGA, and represent the bottleneck in the computation. In our implementation, we use single cycle accumulation, but multi-cycle accumulation is possible to save on hardware resources while scaling up for larger RBM sizes. A fixed point sigmoid function is implemented as a LUT, which allows for speed without expensive hardware operations.  A Python script generates the LUT Verilog code in order to test different bit lengths and fixed point locations.  Finally, a linear feedback shift register (LFSR) generates a pseudo-random number.  The number is compared to the output of the sigmoid LUT and the relevant node is updated with the boolean result.

Results from the bank of visible node registers is buffered to an IO controller with a FIFO.  The IO controller also uses a memory-mapped interface that can program the weights, clamps, and biases.  The controller communicates to our desktop via PCIe. A simple PCIe link is provided through a Xillybus IP Core \cite{Preuer2014ReadyFPGAs} which provides up to 800 MB/s data transfer rate. This is a sufficient speed to get all of the sample data off of the FPGA, but faster speeds are possible for future implementations.  To handle the large data stream from the bus, a C backend was created to serve data to existing Python code for RBM analysis. This C backend is used for analysis of data and to judge the quality of the solution on the FPGA. This FPGA pipeline provides an efficient method for solving the problems of interest, where the limiting factor in computation speed can become the FPGA sampling speed.

\end{subsection}

\end{methods}


\newpage
\bibliographystyle{naturemag}
\bibliography{References/references.bib}

\begin{thebibliography}{10}
\expandafter\ifx\csname url\endcsname\relax
  \def\url#1{\texttt{#1}}\fi
\expandafter\ifx\csname urlprefix\endcsname\relax\def\urlprefix{URL }\fi
\providecommand{\bibinfo}[2]{#2}
\providecommand{\eprint}[2][]{\url{#2}}

\bibitem{Colwell2013TheLaw}
\bibinfo{author}{Colwell, R.}
\newblock \bibinfo{title}{{The chip design game at the end of Moore's law}}.
\newblock In \emph{\bibinfo{booktitle}{2013 IEEE Hot Chips 25 Symposium, HCS
  2013}} (\bibinfo{publisher}{Institute of Electrical and Electronics Engineers
  Inc.}, \bibinfo{year}{2013}).

\bibitem{Waldrop2016MoreMoore}
\bibinfo{author}{Waldrop, M.~M.}
\newblock \bibinfo{title}{{More Than Moore}}.
\newblock \emph{\bibinfo{journal}{Nature}} \textbf{\bibinfo{volume}{530}},
  \bibinfo{pages}{144--147} (\bibinfo{year}{2016}).

\bibitem{Barahona1982OnModels}
\bibinfo{author}{Barahona, F.}
\newblock \bibinfo{title}{{On the computational complexity of ising spin glass
  models}}.
\newblock \emph{\bibinfo{journal}{Journal of Physics A: Mathematical and
  General}} \textbf{\bibinfo{volume}{15}}, \bibinfo{pages}{3241--3253}
  (\bibinfo{year}{1982}).

\bibitem{Kirkpatrick1983OptimizationAnnealing}
\bibinfo{author}{Kirkpatrick, S.}, \bibinfo{author}{Gelatt, C.~D.} \&
  \bibinfo{author}{Vecchi, M.~P.}
\newblock \bibinfo{title}{{Optimization by simulated annealing}}.
\newblock \emph{\bibinfo{journal}{Science}} \textbf{\bibinfo{volume}{220}},
  \bibinfo{pages}{671--680} (\bibinfo{year}{1983}).

\bibitem{Lucas2014IsingProblems}
\bibinfo{author}{Lucas, A.}
\newblock \bibinfo{title}{{Ising formulations of many NP problems}}.
\newblock \emph{\bibinfo{journal}{Frontiers in Physics}}
  \textbf{\bibinfo{volume}{2}}, \bibinfo{pages}{1--14} (\bibinfo{year}{2014}).

\bibitem{Ackley1985AMachines}
\bibinfo{author}{Ackley, D.~H.}, \bibinfo{author}{Hinton, G.~E.} \&
  \bibinfo{author}{Sejnowski, T.~J.}
\newblock \bibinfo{title}{{A learning algorithm for boltzmann machines}}.
\newblock \emph{\bibinfo{journal}{Cognitive Science}}
  \textbf{\bibinfo{volume}{9}}, \bibinfo{pages}{147--169}
  (\bibinfo{year}{1985}).

\bibitem{Korst1989CombinatorialMachine}
\bibinfo{author}{Korst, J.~H.} \& \bibinfo{author}{Aarts, E.~H.}
\newblock \bibinfo{title}{{Combinatorial optimization on a Boltzmann machine}}.
\newblock \emph{\bibinfo{journal}{Journal of Parallel and Distributed
  Computing}} \textbf{\bibinfo{volume}{6}}, \bibinfo{pages}{331--357}
  (\bibinfo{year}{1989}).

\bibitem{Hinton2002TrainingDivergence}
\bibinfo{author}{Hinton, G.~E.}
\newblock \bibinfo{title}{{Training products of experts by minimizing
  contrastive divergence}}.
\newblock \emph{\bibinfo{journal}{Neural Computation}}
  \textbf{\bibinfo{volume}{14}}, \bibinfo{pages}{1771--1800}
  (\bibinfo{year}{2002}).

\bibitem{Tieleman2008TrainingGradient}
\bibinfo{author}{Tieleman, T.}
\newblock \bibinfo{title}{{Training restricted boltzmann machines using
  approximations to the likelihood gradient}}.
\newblock In \emph{\bibinfo{booktitle}{Proceedings of the 25th International
  Conference on Machine Learning}}, \bibinfo{pages}{1064--1071}
  (\bibinfo{publisher}{ACM}, \bibinfo{year}{2008}).

\bibitem{Tieleman2009UsingDivergence}
\bibinfo{author}{Tieleman, T.} \& \bibinfo{author}{Hinton, G.}
\newblock \bibinfo{title}{{Using fast weights to improve persistent contrastive
  divergence}}.
\newblock In \emph{\bibinfo{booktitle}{ACM International Conference Proceeding
  Series}}, vol. \bibinfo{volume}{382}, \bibinfo{pages}{1033--1040}
  (\bibinfo{publisher}{ACM Press}, \bibinfo{year}{2009}).

\bibitem{Geman1987StochasticImages}
\bibinfo{author}{Geman, S.} \& \bibinfo{author}{Geman, D.}
\newblock \bibinfo{title}{{Stochastic Relaxation, Gibbs Distributions, and the
  Bayesian Restoration of Images}}.
\newblock \emph{\bibinfo{journal}{Readings in Computer Vision}}
  \bibinfo{pages}{564--584} (\bibinfo{year}{1987}).

\bibitem{Camsari2017StochasticLogic}
\bibinfo{author}{Camsari, K.~Y.}, \bibinfo{author}{Faria, R.},
  \bibinfo{author}{Sutton, B.~M.} \& \bibinfo{author}{Datta, S.}
\newblock \bibinfo{title}{{Stochastic p -Bits for Invertible Logic}}.
\newblock \emph{\bibinfo{journal}{Physical Review X}}
  \textbf{\bibinfo{volume}{7}}, \bibinfo{pages}{031014} (\bibinfo{year}{2017}).

\bibitem{Jouppi2017In-DatacenterUnit}
\bibinfo{author}{Jouppi, N.~P.} \emph{et~al.}
\newblock \bibinfo{title}{{In-Datacenter Performance Analysis of a Tensor
  Processing Unit}}.
\newblock \emph{\bibinfo{journal}{44th International Symposium on Computer
  Architecture (ISCA),}}  (\bibinfo{year}{2017}).

\bibitem{Ly2009AMachines}
\bibinfo{author}{Ly, D.~L.} \& \bibinfo{author}{Chow, P.}
\newblock \bibinfo{title}{{A high-performance FPGA architecture for Restricted
  Boltzmann Machines}}.
\newblock In \emph{\bibinfo{booktitle}{Proceedings of the 7th ACM SIGDA
  International Symposium on Field-Programmable Gate Arrays, FPGA'09}},
  \bibinfo{pages}{73--82} (\bibinfo{year}{2009}).

\bibitem{Kim2009AImplementation}
\bibinfo{author}{Kim, S.~K.}, \bibinfo{author}{McAfee, L.~C.},
  \bibinfo{author}{McMahon, P.~L.} \& \bibinfo{author}{Olukotun, K.}
\newblock \bibinfo{title}{{A highly scalable restricted Boltzmann machine FPGA
  implementation}}.
\newblock In \emph{\bibinfo{booktitle}{International Conference on Field
  Programmable Logic and Applications}}, \bibinfo{pages}{367--372}
  (\bibinfo{year}{2009}).

\bibitem{Kim2010AMachines}
\bibinfo{author}{Kim, S.~K.}, \bibinfo{author}{McMahon, P.~L.} \&
  \bibinfo{author}{Olukotun, K.}
\newblock \bibinfo{title}{{A large-scale architecture for restricted Boltzmann
  machines}}.
\newblock In \emph{\bibinfo{booktitle}{Proceedings - IEEE Symposium on
  Field-Programmable Custom Computing Machines, FCCM 2010}},
  \bibinfo{pages}{201--208} (\bibinfo{year}{2010}).

\bibitem{Han2016DeepCoding}
\bibinfo{author}{Han, S.}, \bibinfo{author}{Mao, H.} \& \bibinfo{author}{Dally,
  W.~J.}
\newblock \bibinfo{title}{{Deep compression: Compressing deep neural networks
  with pruning, trained quantization and Huffman coding}}.
\newblock \emph{\bibinfo{journal}{4th International Conference on Learning
  Representations, ICLR 2016 - Conference Track Proceedings}}
  (\bibinfo{year}{2016}).

\bibitem{Ullrich2019SoftCompression}
\bibinfo{author}{Ullrich, K.}, \bibinfo{author}{Welling, M.} \&
  \bibinfo{author}{Meeds, E.}
\newblock \bibinfo{title}{{Soft weight-sharing for neural network
  compression}}.
\newblock In \emph{\bibinfo{booktitle}{5th International Conference on Learning
  Representations, ICLR 2017 - Conference Track Proceedings}}
  (\bibinfo{publisher}{International Conference on Learning Representations,
  ICLR}, \bibinfo{year}{2019}).

\bibitem{Chen2015CompressingTrick}
\bibinfo{author}{Chen, W.}, \bibinfo{author}{Wilson, J.~T.},
  \bibinfo{author}{Tyree, S.}, \bibinfo{author}{Weinberger, K.~Q.} \&
  \bibinfo{author}{Chen, Y.}
\newblock \bibinfo{title}{{Compressing neural networks with the hashing
  trick}}.
\newblock In \emph{\bibinfo{booktitle}{32nd International Conference on Machine
  Learning, ICML 2015}}, vol.~\bibinfo{volume}{3}, \bibinfo{pages}{2275--2284}
  (\bibinfo{year}{2015}).

\bibitem{Dally2015High-PerformanceLearning}
\bibinfo{author}{Dally, W.}
\newblock \bibinfo{title}{{High-Performance Hardware for Machine Learning}}.
\newblock In \emph{\bibinfo{booktitle}{Nips 2015}} (\bibinfo{year}{2015}).

\bibitem{Cook1971TheProcedures}
\bibinfo{author}{Cook, S.~A.} \& \bibinfo{author}{A., S.}
\newblock \bibinfo{title}{{The complexity of theorem-proving procedures}}.
\newblock In \emph{\bibinfo{booktitle}{Proceedings of the third annual ACM
  symposium on Theory of computing - STOC '71}}, \bibinfo{pages}{151--158}
  (\bibinfo{publisher}{ACM Press}, \bibinfo{address}{New York, New York, USA},
  \bibinfo{year}{1971}).

\bibitem{Karp1972ReducibilityProblems}
\bibinfo{author}{Karp, R.~M.}
\newblock \bibinfo{title}{{Reducibility among Combinatorial Problems}}.
\newblock In \emph{\bibinfo{booktitle}{Complexity of Computer Computations}},
  \bibinfo{pages}{85--103} (\bibinfo{publisher}{Springer US},
  \bibinfo{year}{1972}).

\bibitem{Hoos2005StochasticSearch}
\bibinfo{author}{Hoos, H.~H.} \& \bibinfo{author}{St{\"{u}}tzle, T.}
\newblock \emph{\bibinfo{title}{{Stochastic Local Search}}}
  (\bibinfo{publisher}{Elsevier}, \bibinfo{year}{2005}).

\bibitem{Ly2008NeuralMachines}
\bibinfo{author}{Ly, D.}, \bibinfo{author}{Paprotski, V.} \&
  \bibinfo{author}{Yen, D.}
\newblock \bibinfo{title}{{Neural Networks on GPUs: Restricted Boltzmann
  Machines}}.
\newblock \bibinfo{type}{Tech. Rep.} \bibinfo{number}{994068682}
  (\bibinfo{year}{2008}).

\bibitem{Li2015AnStreams}
\bibinfo{author}{Li, B.}, \bibinfo{author}{Najafi, M.~H.} \&
  \bibinfo{author}{Lilja, D.~J.}
\newblock \bibinfo{title}{{An FPGA implementation of a Restricted Boltzmann
  Machine classifier using stochastic bit streams}}.
\newblock In \emph{\bibinfo{booktitle}{Proceedings of the International
  Conference on Application-Specific Systems, Architectures and Processors}},
  vol. \bibinfo{volume}{2015-Septe}, \bibinfo{pages}{68--69}
  (\bibinfo{year}{2015}).

\bibitem{Ly2009AMachinesb}
\bibinfo{author}{Ly, D.~L.} \& \bibinfo{author}{Chow, P.}
\newblock \bibinfo{title}{{A multi-FPGA architecture for stochastic Restricted
  Boltzmann Machines}}.
\newblock \emph{\bibinfo{journal}{FPL 09: 19th International Conference on
  Field Programmable Logic and Applications}} \bibinfo{pages}{168--173}
  (\bibinfo{year}{2009}).

\bibitem{Lo2011BuildingMPI}
\bibinfo{author}{Lo, C.} \& \bibinfo{author}{Chow, P.}
\newblock \bibinfo{title}{{Building a multi-FPGA virtualized Restricted
  Boltzmann Machine architecture using embedded MPI}}.
\newblock In \emph{\bibinfo{booktitle}{ACM/SIGDA International Symposium on
  Field Programmable Gate Arrays - FPGA}}, \bibinfo{pages}{189--198}
  (\bibinfo{year}{2011}).

\bibitem{Bremaud1999GibbsSimulation}
\bibinfo{author}{Br{\'{e}}maud, P.}
\newblock \bibinfo{title}{{Gibbs Fields and Monte Carlo Simulation}}.
\newblock In \emph{\bibinfo{booktitle}{Markov Chains}},
  \bibinfo{pages}{253--322} (\bibinfo{publisher}{Springer New York},
  \bibinfo{address}{New York, NY}, \bibinfo{year}{1999}).

\bibitem{Yamaoka2016AAnnealing}
\bibinfo{author}{Yamaoka, M.} \emph{et~al.}
\newblock \bibinfo{title}{{A 20k-spin Ising chip to solve combinatorial
  optimization problems with CMOS annealing}}.
\newblock \emph{\bibinfo{journal}{IEEE Journal of Solid-State Circuits}}
  \textbf{\bibinfo{volume}{51}}, \bibinfo{pages}{303--309}
  (\bibinfo{year}{2016}).

\bibitem{Boyd2018SiliconNews}
\bibinfo{author}{Boyd, J.}
\newblock \bibinfo{title}{{Silicon chip delivers quantum speeds [News]}}.
\newblock \emph{\bibinfo{journal}{IEEE Spectrum}}
  \textbf{\bibinfo{volume}{55}}, \bibinfo{pages}{10--11}
  (\bibinfo{year}{2018}).

\bibitem{Schneider1993AnalogCircuits}
\bibinfo{author}{Schneider, C.~R.} \& \bibinfo{author}{Card, H.~C.}
\newblock \bibinfo{title}{{Analog CMOS Deterministic Boltzmann Circuits}}.
\newblock \emph{\bibinfo{journal}{IEEE Journal of Solid-State Circuits}}
  \textbf{\bibinfo{volume}{28}}, \bibinfo{pages}{907--914}
  (\bibinfo{year}{1993}).

\bibitem{Belletti2009Janus:Computing}
\bibinfo{author}{Belletti, F.} \emph{et~al.}
\newblock \bibinfo{title}{{Janus: An FPGA-based system for high-performance
  scientific computing}}.
\newblock \emph{\bibinfo{journal}{Computing in Science and Engineering}}
  \textbf{\bibinfo{volume}{11}}, \bibinfo{pages}{48--58}
  (\bibinfo{year}{2009}).

\bibitem{Ko2019Flexgibbs:Graphs}
\bibinfo{author}{Ko, G.~G.}, \bibinfo{author}{Chai, Y.},
  \bibinfo{author}{Rutenbar, R.~A.}, \bibinfo{author}{Brooks, D.} \&
  \bibinfo{author}{Wei, G.~Y.}
\newblock \bibinfo{title}{{Flexgibbs: Reconfigurable parallel gibbs sampling
  accelerator for structured graphs}}.
\newblock In \emph{\bibinfo{booktitle}{Proceedings - 27th IEEE International
  Symposium on Field-Programmable Custom Computing Machines, FCCM 2019}},
  \bibinfo{pages}{334} (\bibinfo{publisher}{Institute of Electrical and
  Electronics Engineers Inc.}, \bibinfo{year}{2019}).

\bibitem{Bojnordi2016MemristiveLearning}
\bibinfo{author}{Bojnordi, M.~N.} \& \bibinfo{author}{Ipek, E.}
\newblock \bibinfo{title}{{Memristive boltzmann machine: A hardware accelerator
  for combinatorial optimization and deep learning}}.
\newblock In \emph{\bibinfo{booktitle}{IEEE International Symposium on High
  Performance Computer Architecture}}, \bibinfo{pages}{1--13}
  (\bibinfo{year}{2016}).

\bibitem{Wan2020AModels}
\bibinfo{author}{Wan, W.} \emph{et~al.}
\newblock \bibinfo{title}{{A 74 TMACS/W CMOS-RRAM Neurosynaptic Core with
  Dynamically Reconfigurable Dataflow and In-situ Transposable Weights for
  Probabilistic Graphical Models}}.
\newblock In \emph{\bibinfo{booktitle}{Digest of Technical Papers - IEEE
  International Solid-State Circuits Conference}}, vol.
  \bibinfo{volume}{2020-February}, \bibinfo{pages}{498--500}
  (\bibinfo{publisher}{Institute of Electrical and Electronics Engineers Inc.},
  \bibinfo{year}{2020}).

\bibitem{Dridi2017PrimeGeometry}
\bibinfo{author}{Dridi, R.} \& \bibinfo{author}{Alghassi, H.}
\newblock \bibinfo{title}{{Prime factorization using quantum annealing and
  computational algebraic geometry}}.
\newblock \emph{\bibinfo{journal}{Scientific Reports}}
  \textbf{\bibinfo{volume}{7}}, \bibinfo{pages}{43048} (\bibinfo{year}{2017}).

\bibitem{Wang2013CoherentOscillators}
\bibinfo{author}{Wang, Z.}, \bibinfo{author}{Marandi, A.},
  \bibinfo{author}{Wen, K.}, \bibinfo{author}{Byer, R.~L.} \&
  \bibinfo{author}{Yamamoto, Y.}
\newblock \bibinfo{title}{{Coherent Ising machine based on degenerate optical
  parametric oscillators}}.
\newblock \emph{\bibinfo{journal}{Physical Review A - Atomic, Molecular, and
  Optical Physics}} \textbf{\bibinfo{volume}{88}}, \bibinfo{pages}{063853}
  (\bibinfo{year}{2013}).

\bibitem{McMahon2016AConnections}
\bibinfo{author}{McMahon, P.~L.} \emph{et~al.}
\newblock \bibinfo{title}{{A fully programmable 100-spin coherent Ising machine
  with all-to-all connections}}.
\newblock \emph{\bibinfo{journal}{Science}} \textbf{\bibinfo{volume}{354}},
  \bibinfo{pages}{614--617} (\bibinfo{year}{2016}).

\bibitem{Camsari2017ImplementingMTJ}
\bibinfo{author}{Camsari, K.~Y.}, \bibinfo{author}{Salahuddin, S.} \&
  \bibinfo{author}{Datta, S.}
\newblock \bibinfo{title}{{Implementing p-bits with Embedded MTJ}}.
\newblock \emph{\bibinfo{journal}{IEEE Electron Device Letters}}
  \textbf{\bibinfo{volume}{38}}, \bibinfo{pages}{1767--1770}
  (\bibinfo{year}{2017}).

\bibitem{Borders2019IntegerJunctions}
\bibinfo{author}{Borders, W.~A.} \emph{et~al.}
\newblock \bibinfo{title}{{Integer factorization using stochastic magnetic
  tunnel junctions}}.
\newblock \emph{\bibinfo{journal}{Nature}} \textbf{\bibinfo{volume}{573}},
  \bibinfo{pages}{390--393} (\bibinfo{year}{2019}).

\bibitem{Salakhutdinov2009DeepMachines}
\bibinfo{author}{Salakhutdinov, R.} \& \bibinfo{author}{Hinton, G.}
\newblock \bibinfo{title}{{Deep Boltzmann machines}}.
\newblock \emph{\bibinfo{journal}{Journal of Machine Learning Research}}
  \textbf{\bibinfo{volume}{5}}, \bibinfo{pages}{448--455}
  (\bibinfo{year}{2009}).

\bibitem{Salakhutdinov2010EfficientMachines}
\bibinfo{author}{Salakhutdinov, R.} \& \bibinfo{author}{Larochelle, H.}
\newblock \bibinfo{title}{{Efficient learning of Deep Boltzmann Machines}}.
\newblock In \emph{\bibinfo{booktitle}{Journal of Machine Learning Research}},
  vol.~\bibinfo{volume}{9}, \bibinfo{pages}{693--700} (\bibinfo{year}{2010}).

\bibitem{Savich2011ResourceMNIST}
\bibinfo{author}{Savich, A.~W.} \& \bibinfo{author}{Moussa, M.}
\newblock \bibinfo{title}{{Resource efficient arithmetic effects on RBM neural
  network solution quality using MNIST}}.
\newblock \emph{\bibinfo{journal}{Proceedings - 2011 International Conference
  on Reconfigurable Computing and FPGAs, ReConFig 2011}}
  \bibinfo{pages}{35--40} (\bibinfo{year}{2011}).

\bibitem{Carreira-Perpinan2005OnLearning}
\bibinfo{author}{Carreira-Perpi{\~{n}}{\'{a}}n, M.~Ã.} \&
  \bibinfo{author}{Hinton, G.~E.}
\newblock \bibinfo{title}{{On contrastive divergence learning}}.
\newblock \bibinfo{type}{Tech. Rep.}, \bibinfo{institution}{University of
  Toronto} (\bibinfo{year}{2005}).

\bibitem{Preuer2014ReadyFPGAs}
\bibinfo{author}{Preu{\ss}er, T.~B.} \& \bibinfo{author}{Spallek, R.~G.}
\newblock \bibinfo{title}{{Ready PCIe data streaming solutions for FPGAs}}.
\newblock In \emph{\bibinfo{booktitle}{Conference Digest - 24th International
  Conference on Field Programmable Logic and Applications, FPL 2014}}
  (\bibinfo{publisher}{Institute of Electrical and Electronics Engineers Inc.},
  \bibinfo{year}{2014}).

\end{thebibliography}


\newpage
\begin{addendum}

\item [Acknowledgements] 
This work was supported by ASCENT, one of six centers in JUMP, a Semiconductor Research Corporation (SRC) program sponsored by DARPA.

\item [Author Contributions] 
Model Synthesis and Analysis was performed by S.P.; FPGA Programming was performed by S.P and P.C.; Manuscript was co-wrote by S.P, P.C and S.S; S.S supervised the research. 
All authors contributed to discussions and commented on the manuscript. 

\item [Competing Interests] 
The authors declare that they have no competing financial interests.

\item [Correspondence] 
Correspondence and requests for materials can be addressed to either S.P. \\ (saavan@berkeley.edu) or S.S. (sayeef@berkeley.edu).

\clearpage
\begin{figure*}
\begin{centering}
\includegraphics[width=0.88\linewidth]{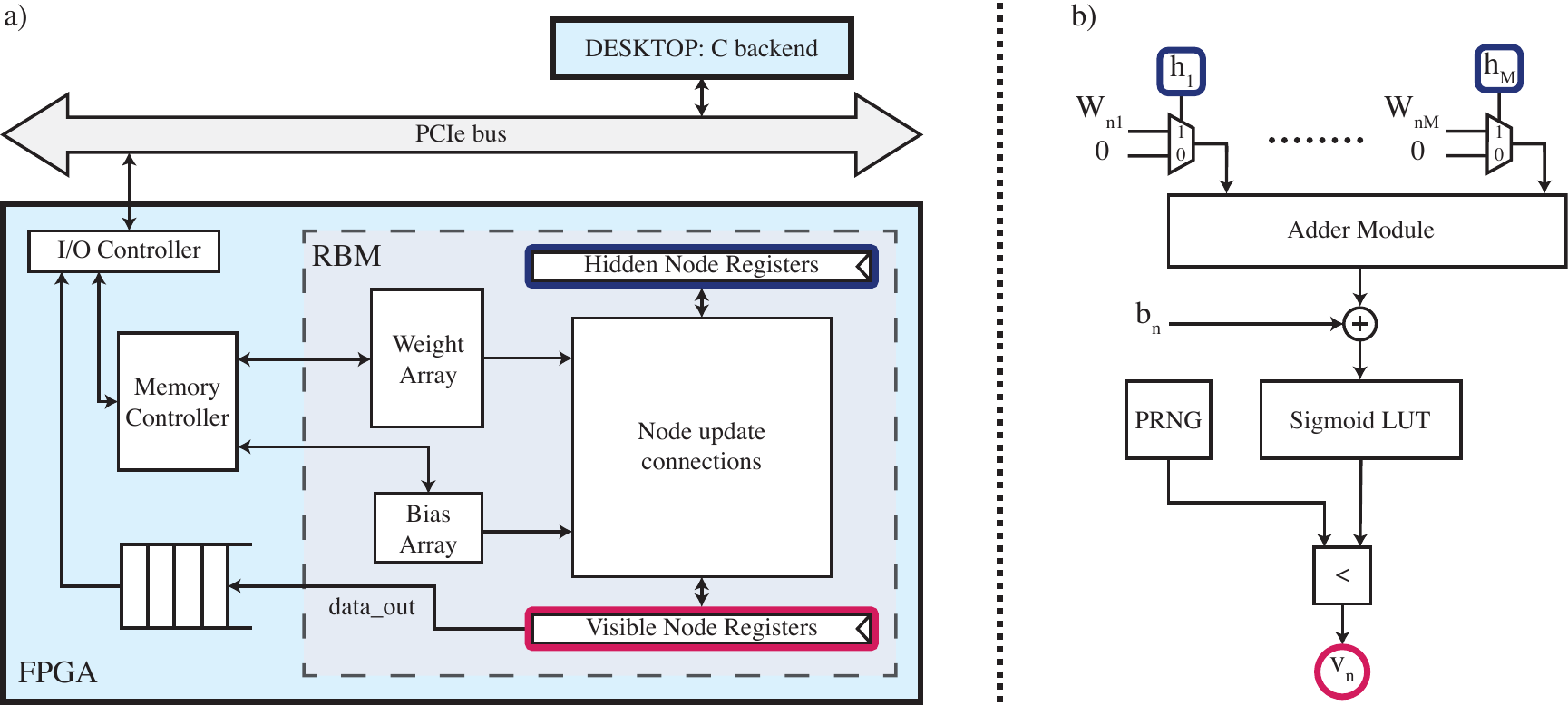}
\par\end{centering}
\caption{\label{fig:fpga} (Supplementary) \textbf{FPGA Architecture} {\bf (A)\/} Memory and compute hierarchy.  The RBM consists of memory to hold the weight, bias, and clamp values, registers to hold the node values, and circuitry to perform the node updates, which take up the bulk of the resources.  The output is buffered to the IO controller that communicates results to the PCIe bus. A C backend reads in the data stream from PCIe, and can program the weights and biases from the memory controller.
{\bf (B)\/} Example of a node update connection. Given $M$ hidden nodes, the figure depicts the circuitry to update the $n^{th}$ visible node. The hidden nodes binary mask the $n^{th}$ row of the weight matrix.  The results are accumulated in the adder module and added to the $n^{th}$ visible bias.  It is then passed through a sigmoid LUT and compared to the output of a PRNG to update the value of the visible node. }
\end{figure*}

\clearpage
\begin{table*}

\centering
\begin{tabular}{ |p{2.5cm}||p{2.5cm}|p{2.5cm}|p{2.5cm}| p{2.5cm}| p{2.5cm}| } 
 \hline
 \thead{RBM Size \\ (Vis x Hid)} & \thead{LUT Usage \\ (Absolute)} & \thead{LUT Usage \\ (\%)} & \thead{FF Usage \\ (Absolute)} & \thead{FF Usage \\ (\%)} & \thead{Power \\ (W)} \\
 \hline\hline
 8x32 & 14811 & 1.25 & 15437 & 0.65 & 5.2\\  \hline
 16x64 & 33564 & 2.84  & 23717 & 1.00 & 5.2 \\  \hline
 32x128 & 117931 & 9.98 & 52582 & 2.22 & 5.5\\  \hline
 64x256 & 418694 & 35.42 & 159428 & 6.74 & 5.6\\ \hline
 64x512 & 736800 & 62.32 & 270182 & 11.43 & 5.9 \\ \hline
 80x600 & 990417 & 83.77 & 431544 & 18.25 & 6.2 \\
 \hline
\end{tabular}
\caption{\label{tab:utilization} (Supplementary) \textbf{FPGA Utilization Utilization numbers for FPGA and various RBM sizes}  
All usage numbers reported are for 8 bit weights and biases . The usage shows that the FPGA is not memory limited for the problem sizes we are interested in, but compute limited, as the LUT usage goes up much faster than the FF usage as the problem size grows. All weights and biases fit in on chip SRAM, allowing for fast access and data reuse.}
\end{table*}

\clearpage
\begin{table*}

\centering
\begin{tabular}{ |p{5cm}||p{3cm}|p{3cm}|p{3cm}| p{3cm}|  } 
 \hline
 \thead{Model} & \thead{Visible Units} &  \thead{Hidden Units} & \thead{Training Time \\ (minutes)} \\ 
 \hline\hline
1 bit Adder     & 5 & 6  & 1\\ \hline
2 bit Adder     & 8 & 28 & 13.5 \\  \hline
4 bit Adder     & 14 & 64 &  133 \\  \hline
8 bit Adder     & 26 & 96 & 201  \\  \hline
16 bit Adder    & 50 & 128 & 321 \\ \hline
32 bit Adder    & 98 & 192 & 13000 (approx.) \\ \hline
4 bit Multiplier & 8 & 16 & 46 \\ \hline
6 bit Multiplier & 12 & 48 & 79 \\ \hline
8 bit Multiplier & 16 & 64 & 655 \\ \hline
10 bit Multiplier & 20 & 144 & 4063 \\ \hline
12 bit Multiplier (Merged) & 60 & 352 & 512 \\ \hline
16 bit Multiplier & 32 & 512 & 3611 \\  \hline
16 bit Multiplier (Merged) & 78 & 576 & 5156 \\
\hline
\end{tabular}
\caption{\label{tab:models} (Supplementary) \textbf{Model sizes and training times for various RBMs} As the size of the RBM grows, both the training time for model convergence and the number of hidden units needed to model the distribution both increase. We find that the largest multiplier model that can be fit on our version of the FPGA design is the 16 Bit Multiplier/Factorizer. Model analysis and training for models larger than the 16 bit multiplier took too long to train, and are not displayed here.}
\end{table*}

\end{addendum}

\end{document}